\documentclass[journal]{IEEEtran}
\usepackage{cite}
\usepackage{amsmath,amssymb,amsfonts}
\usepackage{algorithmic}
\usepackage{graphicx}
\usepackage{textcomp}
\usepackage{xcolor}
\def\BibTeX{{\rm B\kern-.05em{\sc i\kern-.025em b}\kern-.08em
    T\kern-.1667em\lower.7ex\hbox{E}\kern-.125emX}}
\usepackage{ragged2e}
\usepackage{booktabs} 
\usepackage{hyperref}
\usepackage{cleveref}
\usepackage{multirow}
\usepackage{color}
\usepackage{diagbox}
\usepackage[dvipsnames]{colortbl}
\usepackage{amssymb}
\usepackage{balance}
\usepackage{paralist}
\usepackage{enumitem}
\usepackage{times}
\usepackage{latexsym}
\usepackage{array}
\usepackage{etoolbox}
\usepackage{subfigure}
\usepackage{diagbox}
\usepackage{amssymb}
\usepackage{balance}
\usepackage{fixltx2e}
\usepackage{xspace}

\usepackage{amsfonts}
\usepackage{wrapfig}
\usepackage{graphicx}
\usepackage{amsmath} 

\newcommand{\eg}{e.\,g.\,, }

\newcommand{\cf}{{cf.\,}}
\DeclareMathOperator{\ave}{average} 

\usepackage{soul}
\definecolor{greenish}{rgb}{0.0,0.7,0.0}

\newcommand{\close}{\textsc{Close-CAR\,}}
\newcommand{\mix}{\textsc{Mix-CAR\,}}
\newcommand{\mpart}{\textsc{MuSe-CAR-Part\,}}
\newcommand{\gocart}{\textsc{go-CaRD\,}}

\newcommand{\cl}{\textsc{close\,}}
\newcommand{\mi}{\textsc{mix\,}}
\newcommand{\mpartshort}{\textsc{mpart\,}}

\newcommand{\dn}{\textsc{DenseNet201\,}}
\newcommand{\irn}{\textsc{InceptionResNetV2\,}}
\newcommand{\iv}{\textsc{InceptionV3\,}}

\newcommand{\ivall}{\textsc{Inception\,}}
\newcommand{\ivfour}{\textsc{InceptionV4\,}}
\newcommand{\mnv}{\textsc{MobileNetV2\,}}
\newcommand{\mnl}{\textsc{NASNetLarge\,}}
\newcommand{\mnm}{\textsc{NASNetMobile\,}}
\newcommand{\rn}{\textsc{ResNet50\,}}
\newcommand{\vs}{\textsc{VGG16\,}}
\newcommand{\vn}{\textsc{VGG19\,}}
\newcommand{\xce}{\textsc{Xception\,}}

\newcommand{\darknet}{\textsc{Darknet\,}}
\newcommand{\dtiny}{\textsc{TinyDarknet\,}}
\newcommand{\squee}{\textsc{SqueezeNet\,}}

\usepackage[activate]{microtype}
\begin{document}

\title{Domain Adaptation with Joint Learning for \\Generic, Optical Car Part Recognition and Detection Systems (Go-CaRD)} 

\author{ Lukas Stappen,~\IEEEmembership{Member~IEEE,}
        Xinchen Du,
        Vincent Karas,~\IEEEmembership{Member~IEEE,}
        Stefan M\"uller,
        and~Bj{\"o}rn W.\ Schuller,~\IEEEmembership{Fellow IEEE, ISCA}

}

%
%

\markboth{under review}
{Shell \MakeLowercase{\textit{et al.}}: Bare Demo of IEEEtran.cls for IEEE Journals}



\maketitle

\begin{abstract}
Systems for the automatic recognition and detection of automotive parts are crucial in several emerging research areas in the development of intelligent vehicles. They enable, for example, the detection and modelling of interactions between human and the vehicle. In this paper, we quantitatively and qualitatively explore the efficacy of deep learning architectures for the classification and localisation of 29 interior and exterior vehicle regions on three novel datasets. Furthermore, we experiment with joint and transfer learning approaches across datasets and point out potential applications of our systems. Our best network architecture achieves an F1 score of 93.67\,\% for recognition, while our best localisation approach utilising state-of-the-art backbone networks achieve a mAP of 63.01\,\% for detection. 
The \mpart dataset, which is based on a large variety of human-car interactions in videos, the weights of the best models, and the code is publicly available to academic parties for benchmarking and future research\footnote{Demonstration and instructions to obtain data and models: https://github.com/lstappen/GoCarD.}.


\end{abstract}

\begin{IEEEkeywords}
car parts; object detection; domain adaption; visual perception; intelligent vehicles
\end{IEEEkeywords}

%
\IEEEpeerreviewmaketitle

\section{INTRODUCTION}
In the development of intelligent vehicles, there is a strong need for robust models that can automatically detect the parts and regions of cars, both inside and outside the vehicle. Opening doors on nearby vehicles, for example, must be identified to avoid collisions~\cite{zhu2019novel, liu2016radar}, while the autonomous transition shifts the focus towards the interior, necessitate a holistic understanding of the cabin for intelligent personal assistants and advanced user interactions~\cite{9220832, 6145755, vogel2018emotion}. Part recognition can also be used in production environments for automatic quality control~\cite{Luckow.2016b, MurielMazzetto.2020}.

Developing methods for automatic object detection in images and videos is the major goal of computer vision. Reliable approaches are particular challenging in domains in which objects of the same class appear in a wide range of variations and environments. In practise, this leads often to datasets which cover very similar or the same classes, but serving different purposes. A generic perception system must be able to recognise parts and regions across the non-trivial automotive domain; a single car model can have more than $10^{24}$ variations~\cite{pil2004linking} due to the vast number of equipment options. Additionally, in real-world scenarios, such a system must be able to operate in different, changing, and subpar perspectives and lighting conditions, robust against human occlusion, and with images of varying quality, to be assimilable into more complex frameworks. 
However, to date, literature on generic optical detection systems that meet or even consider all these criteria exists is lacking.
Typically, investigations focus on car parts in a static, lab-like environment without significant human interactions. Existing methods designed for fine-grained recognition of vehicles often only function in fixed lighting conditions and using images from specific viewpoints, such as frontal and rear view images~\cite{ hu2015learning}, or extract distinctive parts automatically, but cannot assign a label to them~\cite{simon2015neural}.

A statistical system that detects regions of car exteriors in 110 images was developed by~\cite{chavez2011vision}, while~\cite{Luckow.2016b} proposes a visual inspection system for the manufacturing process. The latter recognises different vehicle properties related to quality faults of four vehicle models, using AlexNet, GoogleNet, and Inception V3, in 82\,000 images from production environments and 106 in-the-wild images from Twitter. The system achieved an F1 score of 87.2\,\% on the top five classes. Studies based on the Stanford Cars and BMW-10 dataset~\cite{krause20133d, sharif2017framework} predict the make, model and year, but not the underlying vehicle topology, while investigations using the BoxCars focus on re-identification in traffic surveillance~\cite{sochor2016boxcars, wang2017orientation}.

Addressing the gap in the literature, the contributions of this work are threefold:
\begin{itemize}
  \item We provide a broad empirical comparison of neural network models, derived from state-of-the-art computer vision architectures, for automotive part recognition and detection.
  \item To do so, we created and labelled more than 12\,000 car part images in challenging real-life environments, from which we make a share available to other researchers$^1$.
  \item Additionally, we demonstrate an efficient domain adaption, joint learning procedure for the task of car part detection. We show that this procedure is more robust than other approaches, such as unsupervised domain adaptation and fine-tuning. 
\end{itemize}

In this work and in line with computer vision literature \cite{voulodimos2018deep}, we define the task of recognition as identifying the main object in an image, hence, every image belongs to a single class. Detection differs insofar as several objects have to be identified as well as localised, resulting in a bounding box around each object of interest. 

Most deep learning detection networks aimed at solving complex vision tasks rely on general-purpose image detection systems, trained on large labelled datasets (e.g. ImageNet) \cite{tan2018survey, krizhevsky2017imagenet}. 
The transfer of knowledge, meaning taking trained network weights from a general domain and fine-tuning them to a specific domain, is called transfer learning. 
It has been found to be more efficient than training Convolutional Neural Networks (CNN) from scratch or using off-the-shelf pre-trained CNN features \cite{sun2016return}. In this work, we focus on domain adaption, a sub-field of transfer learning, using representations learnt by a CNN while solving a problem in a source domain, to a different, but closely related target domain \cite{sun2015survey}. The assumption is, that the source and target domains share the same feature space (but different distributions). Here, we train a network using a large dataset depicting no humans, while injecting images from a small dataset with human obstruction -- due to interaction with the parts -- into the training set to predict the latter.


The remainder of this paper is organised as follows. We provide an overview of the theoretical deep neural network mechanisms and convolutional architectures we base our developments on in \Cref{sec:method}. In~\Cref{sec:set}, we introduce three new datasets with distinctive characteristics suitable for the tasks: \close (\cl) consisting of close shot images under sub-optimal conditions, \mix (\mi) a multi-label dataset covering 18 different BMW models of the last six years, each with up to a hundred equipment options. Both cannot be distributed for licensing reasons. However, we also introduce \mpart (\mpartshort) capturing elevated human vehicle interactions in real-life videos to support future research by academia in this field. Next, we explain modifications to the architectures, our networks are based on, and chosen experimental settings, including in- and out-of-domain transfer capabilities, for each experiment in~\Cref{sec:experiments}. We discuss the quantitative and qualitative results obtained using these approaches in~\Cref{sec:results} and conclude by suggesting potential applications in~\Cref{sec:app}. On the test set, our best-performing system achieves an F1 of $93.76$\,\% (fine-tuned \rn) in a single label setting and a mean average precision (mAP) of $63.01$\,\% on \mi and $41.07$\,\% on \mpartshort when jointly trained for detection - outperforming unsupervised adaption and fine-tuning approaches.

\section{METHODOLOGY: DEEP NEURAL NETWORKS FOR OPTICAL RECOGNITION AND DETECTION}\label{sec:method}

\subsection{Convolutional Neural Networks}
Computer vision is a field concerned with developing algorithms that can detect objects in images in a manner similar to humans. Deep neural networks, extracting high-level features across a large number of layers, form the state-of-the-art in computer vision. In the following section, we give an overview of the most integral components of our trained models -- CNNs -- along with its mathematical equations \cite{krizhevsky2017imagenet}.

Consider an image source $I$ with the dimensions: the size of the height $n_H$, the size of the width $n_W$, and the number of channels $n_C$. Usually RGB images are used, setting $n_C$ to 3 and later to the number of filters per layer $l$.
In the convolutional operation, we train filters $K$, having odd, squared dimensions as a property, which enables equal sized surroundings of the pixels.
Given the image and filter, a convolutional product $\operatorname{CONV}$ is carried out between these two, where each matrix element is the sum of the elementwise multiplication:

\begin{equation}
\operatorname{CV_{op}}(I, K)_{x, y}=\sum_{i=1}^{n_{H}} \sum_{j=1}^{n_{W}} \sum_{k=1}^{n_{C}} K_{i, j, k} I_{x+i-1, y+j-1, k},
\end{equation}

resulting in the dimensions 

\begin{multline}
d = \left(\left\lfloor\frac{n_{H}+2 p-f}{s}+1\right\rfloor,\left\lfloor\frac{n_{W}+2 p-f}{s}+1\right\rfloor\right); \\ 
s>0,
\end{multline}

where $f$ is the dimension of the filter; $s$ indicates the stride step taken; $p$ defines the type of padding, so that a valid convolution which does not use padding leads to $p = 0$ and a same convolution pads the input matrix to ensure that the output will have the same shape ($p=\frac{f-1}{2}$). A small stride increase the size of the output and vice-versa. The convolutional step is often followed by a pooling step, which downsamples the number of image features through summation. The two most common pooling operations are $average$ pooling, where all elements of the filter are averaged and $max$ pooling where only the maximum value is returned. 

These operations a combined to layers. A convolutional layer applies the convolutional operation with many trainable filters $\left(f^{[l]} \times f^{[l]} \times n_{C}^{[l-1]}\right) \times n_{C}^{[l]}$ and a broadcasted bias $b = (1 \times 1 \times 1) \times n_{C}^{[l]]}$ followed by, an often non-linear, activation function $\sigma$: 

\begin{equation}
\operatorname{CV_L}\left(a^{[l-1]}, K^{(n)}\right)_{x, y}=\sigma^{[l]}\left(\operatorname{CV_{op}}(a^{[l-1]}, K^{(n)}) + b_{n}^{[l]}\right),
\end{equation}
where $a^{[0]}$ being the input image. The parameterless pooling layer can be formalised as

\begin{multline}
\operatorname{POOL}\left(a^{[l-1]}\right)_{x, y, z}=\lambda^{[l]}\left(a_{x+i-1, y+j-1, z}^{[l-1]}\right); \\ {(i, j) \in\left[1,2, \ldots, f^{[l]}\right]^{2}},\,\,\,
\end{multline}

where $\lambda$ is the pooling function ($\ave$ or $\max$). These layers are combined in various ways to form blocks, which are multiple times repeated for feature extraction. When feature extraction by the CNN blocks is completed, the features are flatten and passed to a fully connected feedforward layer. As before, an input vector $a$ is taken and transformed to an output vector $z$:

\begin{equation}
z^{[i]}=\sum_{l=1}^{n_{i-1}} w_{l}^{[i]} a_{l}^{[i-1]}+b^{[i]},
\end{equation}

where $w_l$, $b_l$ are the layer weights and bias, respectively. Usually, a feedforward layer has a differentiable, non-linear activation function $\psi$ (\eg ReLU), so that the output transforms to the input of the next layer $a^{[i]}=\psi^{[i]}\left(z^{[i]}\right)$. If the last layer predicts the target $y$, a sigmoid or softmax function is often chosen for $\psi$.

An objective function $\mathcal{L}$ calculates the prediction error between the real target $y$ and the predicted $\hat{y}$ and propagates it back through the network by a gradient decent optimisier. In this process, the network parameters $\theta$ are adjusted with the goal to reduce the error and improve the prediction result. 
\begin{equation}
J(\theta)=\frac{1}{m} \sum_{i=1}^{m} \mathcal{L}\left(\hat{y}_{i}, y_{i}\right).
\end{equation}

Most more advanced CNN architectures combine and tweak the previous introduced layers. Following, we introduce two of the most common network blocks. A residual block \cite{he2016deep} injects residuals from earlier layers, \eg $n-2$, basically skipping connections to destabilise the output:
\begin{equation}
a^{[i]}=\psi^{[i]}\left(z^{[i]}+\mathbf{W}_{\mathbf{s}} a^{[i-2]}\right),
\end{equation}
where $\mathbf{W}$ is a mapping matrix if $a^{[i]}$ and $a^{[i-2]}$ have different shapes. This idea enables consistent training of very deep networks. 

Inception blocks are also widely used~\cite{szegedy2017inception,chollet2017xception}. They feed the input into multiple convolutional (and pooling) layers in parallel, each with different filter sizes $n_C$. This is followed by the concatenation of the output filters, which help to approximate an optimal local sparse structure.

Next, we briefly introduce a number of popular recognition and detection architectures, which are based on these network mechanisms. 

\subsection{Recognition}
In general, a visual recognition systems can be represented by a function $y = F(I)$, which maps (classifies) an image $I$ to a class label $y$ of the displayed object instance. Simonyan and Zisserman~\cite{simonyan2014very} introduced 16 and 19 layers CNNs, \vs and \vn, respectively,  which significantly pushed the benchmark on the 2014 ImageNet challenge.
\rn~\cite{he2016deep} employed residual connections to train networks that were much deeper than the VGG nets, leading to a further increase in performance. 
\ivall architectures~\cite{szegedy2016rethinking} are based on the idea to use wide networks performing multiple convolutions in parallel, instead of ever deeper networks. \iv is an evolution, adding regularisation and batch normalisation to the auxiliary classifiers and applying label smoothing.  
Updating the \iv network,~\cite{szegedy2017inception} introduced \irn that employs residual connections in the inception modules, which accelerates the training process. Its processing cost is similar to the non-residual \ivfour, introduced in the same work.
Inspired by \iv, \xce~\cite{chollet2017xception} was developed, replacing the Inception modules with depth-wise separable convolutions, i.\,e., convolutions that act on different channels of the previous layer's output. \xce has a similar number of parameters as \iv, but it is simpler to implement and showed an increase in performance on several benchmarks.
Compared to other approaches, \dn~\cite{huang2017densely} is a CNN, in which each layer has a feed-forward connection to each other layer, not just its immediate successor. This helps the network propagate features and combat the vanishing gradient problem (an issue leading to 0, so called dead, neurons and occurs occasionally when training very deep networks), as well as decreasing computational cost due to a reduced number of parameters. 

\mnv~\cite{sandler2018mobilenetv2} was developed to reduce the memory footprint making deep neural networks more suitable for mobile applications. Its architecture builds upon residual connections between linear bottleneck layers. 
\mnl and \mnm~\cite{zoph2018learning} follow a different concept of optimisation, searching for a building block on a small dataset, followed by a transfer of stacked blocks to a larger dataset. 
Most of these architectures require specific image pre-processing.
\subsection{Detection}
In contrast to recognition, a visual object detection system is often learnt as a regression task, where also the position of an object on a given image $x$ is predicted aside of the class. This can be expressed by a predicted object ($y$) with the properties: central coordinates X ($y_X$) and Y ($y_Y$), the height ($y_H$) and width ($y_W$) of the bounding box, as well as the confidence of the class ($y_{C}$).

\subsubsection{Framework}
``You-Only-Look-Once`` (YOLO V3) is one of the most efficient and popular object detection frameworks~\cite{redmon2018yolov3}. Unlike similar algorithms, all classes and bounding boxes are predicted simultaneously. Learning classes in dependence of each other leads to a performance advantage and increases context image understanding. So is a video frame extracted from a video stream and divided into an $S \times S$ grid. Each grid cell has a feature vector with the size of \textit{the number of anchors} *\textit{a 5-dimensional object vector} + \textit{the number of classes}. 
The so-called anchors (width-height pairs) enable the network to detect and predict parallel objects of different sizes equally efficiently.
\subsubsection{Backbones}
For the actual prediction functionality, a so-called backbone has to be developed and trained, which corresponds to a neural network. We choose two different backbones~\cite{redmon2018yolov3}, derived from the same neural network blocks
\darknet and for resource efficient usage (\eg mobile applications), the parameter reduced network \dtiny. As almost all applications benefit from a resource-efficient implementation and storage, we also adjust and train a \squee, which reduces parameters by downsampling and a `squeeze and extend' process containing a fire module with decreased filter size and input channel number. A detailed description to the architectural mechanisms of this network type can be found in~\cite{iandola2016squeezenet}.

\begin{figure}[t!] 
\begin{minipage}{\columnwidth}\centering 
  \includegraphics[width=\textwidth]{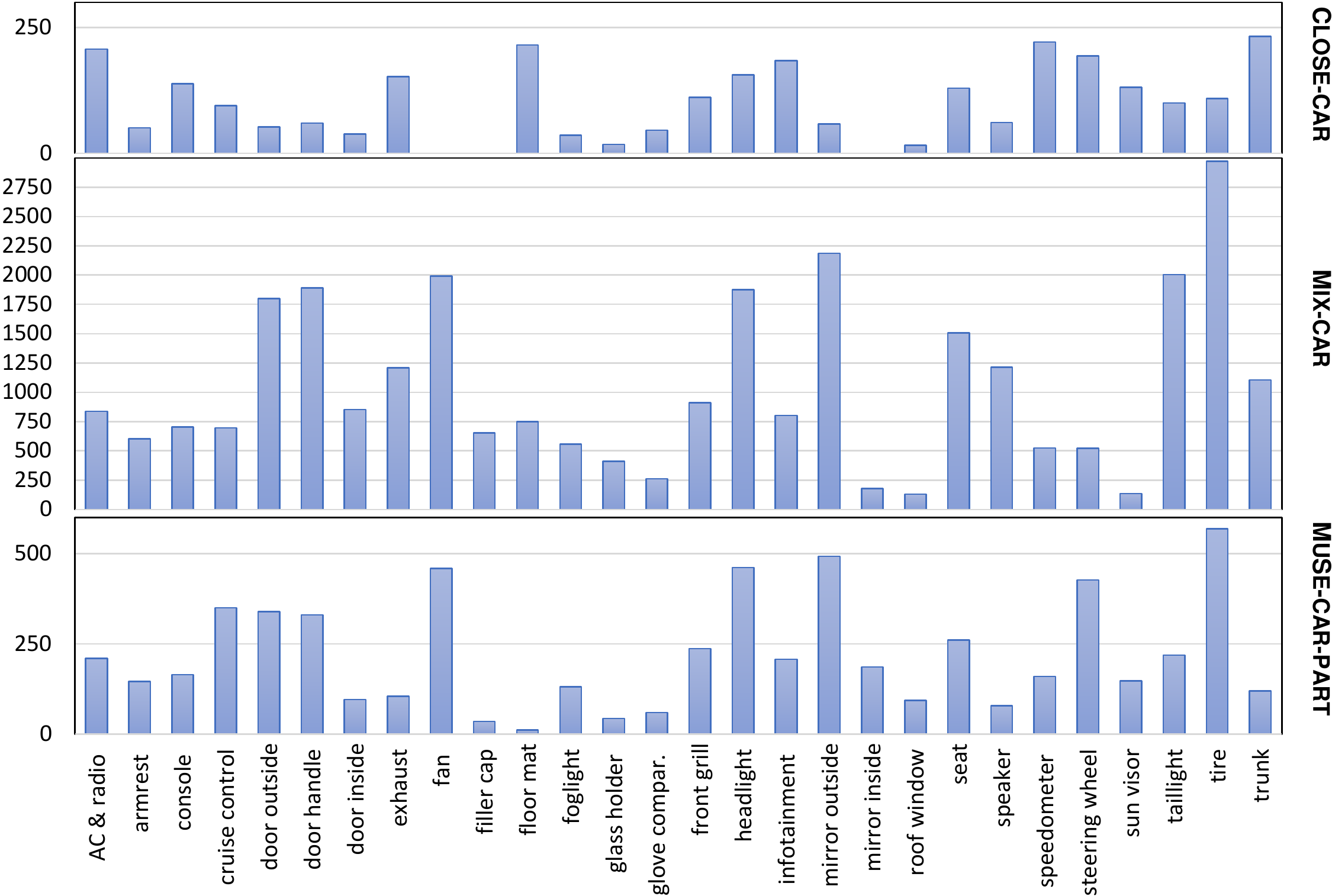} 
 \end{minipage}
  \caption{Label distribution of \close, \mix, and \mpart. \close has slightly different classes, so that there are no labels for the "fan", "filler cap" and "mirror inside" classes. For visualisation reasons, the two classes capturing infotainment systems are combined to one. The same logic was applied to classes including radio. }
  \label{fig:distribution} 
\end{figure}
\subsection{Transfer learning and domain adaption}
Consider two data distributions $I_1$ and $I_2$ from different domains $D_1$ and $D_2$. Transfer learning assumes that both are still similar enough to each other that initialising $F_{w=I_1}(I_2)$, \eg a CNN, with trained weights based on $I_1$, leads to a more efficient training process on $I_2$ compared to a random initialising and a solely training on $D_2$ \cite{lu2015transfer}. After initialisation, the network can either be used as a pure feature extractor without any further training, partially fine-tuned, or fully retrained. When partially fine-tuned, it is common to either freeze weights of lower network layers or freeze the entire pre-initialised base and add a head on top, consisting of a few trainable layers.

Supervised domain adaption assumes that $I_1$ and $I_2$ are from a strongly related domain $D_{1\times2}$ but have different distributions \cite{sun2016return, ben2010theory}, for example, cars and trucks are both vehicles with common features (doors, tiers, etc.) but with different functionality (\eg passenger vs. good transport). In contrast to transfer learning, one network is not fully trained on one dataset and then fine-tuned on another. Instead a share of (labelled or unlabelled) data $X$ from the source domain is injected into the training's set of the target domain, thus, jointly trained. Therefore training $F_2(I_2 + X\,\% \times I_1)$ from samples coming from both distributions.

\subsection{Measures}
In classification tasks, measures that do not take class imbalance into account, \eg accuracy, might lead to a misinterpretation of a model's performance. To avoid any sort of bias due to imbalanced classes, we aim for a balance between recall and precision. We opt for the F1 score (macro), the harmonic mean of recall and precision, to report all our classification results. It is defined by:

\begin{equation}
    F1\,score = \frac{2\cdot TP}{2\cdot TP + FP + FN},
\end{equation}
where TP are the true positive, FP the false positive and FN the false negative predictions on a classical confusion matrix.

For numerical performance comparison of the detection tasks, we used mAP, which is based on the Intersection of Union (IoU) reflecting the area under the interpolated precision-recall curve averaged across all unique recall levels.  IoU divides the intersection area $I$ by the union area $U$ of the prediction and ground truth bounding boxes:
\begin{equation}
I o U=\frac{|I|}{|U|}=\frac{|A \cap B|}{|A \cup B|}.
\end{equation}
The resulting measure is compared to a pre-defined threshold. The threshold enables to categorise the prediction into a confusion matrix from which the Average Precision (AP) is derived. The true positives are considered if the correct class is predicted and the IoU is larger than a threshold.

\section{DATASET AND PREPARATION}\label{sec:set}

In this section, we introduce three separately collected and annotated real-world datasets. In each of them, vehicle parts appear in different environments and conditions, giving them unique characteristics. \Cref{fig:distribution} depicts an overview of the number of labels per class for each dataset. As expected by the nature of the datasets, they are unevenly distributed across the classes. 

\subsection{Close-CAR}
\begin{figure}[t!] 
\begin{minipage}{\columnwidth}\centering 
  \includegraphics[width=\textwidth]{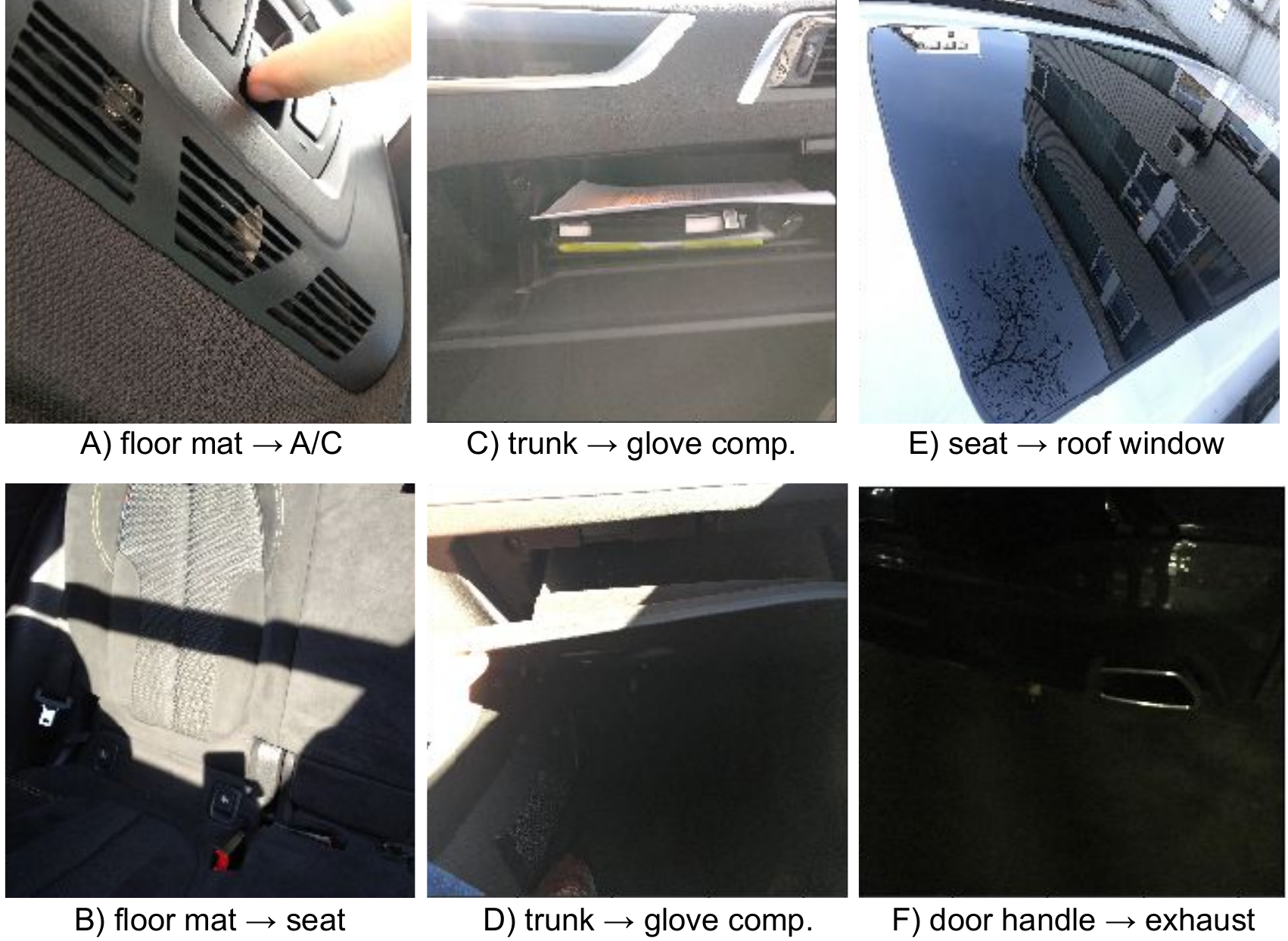} 
 \end{minipage}
  \caption{Examples of incorrectly predicted images (predicted $\rightarrow$ real class) by our model architecture on the test set of \close.}
  \label{fig:fullshot} 
\end{figure}

\close consists of $2\,809$ real-world, close-shot images of two kinds: 1\,743 images of interior parts comprising 19 classes, and 1\,066 exterior parts comprising 10 classes\footnote{interior classes: A/C, A/C infotainment, A/C radio infotainment, A/C radio, armrest, console, cruise control, door inside, floor mats, glass holder, glove compartment, infotainment, radio, roof window, seat, speaker, speedometer, steering wheel, and sun visor; exterior classes: door ex, door handle, exhaust, foglight, grills, headlight, mirror ex, taillight, tire, and trunk}. Every image depicts only a single car part\footnote{or a combination of up to four \eg for A/C, radio and infotainment which are physically located side by side} of various car makes, types (such as SUVs or sedans) and models. Resolution and capture angle to the object vary across the images, which were taken with several hand-held devices, such as the iPhone 5 and 6. Since the photographs were taken under real-world conditions, many suffer from overexposure, underexposure, blurring, reflections from metallic surfaces and shadows (\cf \Cref{fig:fullshot}). This makes the dataset challenging for generic recognition. Training, development and test sets were partitioned in a class-stratified 80\,\%-10\,\%-10\,\% split. We consider this as a brand, model and environment independent dataset.

\subsection{Mix-CAR}
\mix is a multi-label, multi-class real-world dataset that contains 15\,003 images of cars from 18 BMW models, each with up to 100 different cars and options. During and before the production start of a newly developed car model, manufacturers produce a large number of them in various equipment combinations\footnote{\eg various styles of painting colours, upholstery/interior trims, wheels, seats, (head-up) display, towbar, loudspeaker and ventilation covers etc.}. These vehicles are only intended for internal use. They ensure that minor, for example optical and non-safety-related quality issues of the novel parts are found when used in an everyday context, and eliminated, before the start of large-scale production. In the context of these customer-oriented tests, the vehicles are photographically documented before they are lent out to the test person. Including a large number of equipment variations; they enable robust discriminative features to be learnt. Robustness is of particular importance, as cars have an extremely high number of potential product variants, \eg a Mercedes E-class total variations surpassed the order of $10^{24}$~\cite{pil2004linking}, resulting a high variance in images.
The dataset depicts the cars' interior and exterior. We identified 8\,113 of the images (4\,724 exterior and 3\,389 interior) as suitable for labelling distributed on 29 car parts. Each image averages 3.7 bounding boxes with 1 to 15 unique labels. \Cref{fig:distribution} shows that parts were labelled more often that are present more than once on a car, such as tyres (4) and lights (one for each side). We follow the same partition logic as for \cl. 
%
\subsection{MuSe-CAR-Part} 
The previous dataset, guarantees a high variance of optical features within a model, however, could lead to a bias towards BMW-typical, optical features. \mpart is a subset of the multimodal in-the-wild dataset MuSe-CAR, originally collected from YouTube to study multimodal sentiment analysis in-the-wild. The 300 videos provide complex in-the-wild footage, including a range of shot sizes (close-up, medium, and long), camera motion (free, stable, unstable, zoom, and fixed), moving objects, highly variety in backgrounds, and people interacting with the car they are reviewing (\cf \Cref{fig:yoloinsights}). We selected 74 videos from 25 different channels and sampled 1\,124 frames across several topic segments. A detailed description of the properties of the original dataset can be found in
~\cite{stappen2020muse, stappen2021multimodal}. In total, 29 classes were labelled according to \mi, resulting in 6\,146 labels averaging 5.47 labels per frame. From \Cref{fig:distribution}, we can see that there are similar patterns in the label distribution to \mix, for example, fans, headlights, wing mirrors, tyres are disproportionately often represented. In contrast, other classes have very few labels, such as floor mats, glass holders and glove compartments. The dataset is available to academic institutions for further research.


\section{Go-CaRD EXPERIMENTS}\label{sec:experiments}
The conducted experiments are implemented in phython using the deep learning framework Tensorflow and run on four Tesla V100 GPUs (128\,GB GPU RAM in total).
\subsection{Weight transfer and joint cross-corpus learning}

To train models on our limited recognition dataset efficiently, we use out-of-domain transfer-learning. Initialising the network using weights previously trained on the large, general purpose dataset ImageNet~\cite{deng2009imagenet} to stabilise the training of the low-level filters. For the detection task, we apply inner-domain transfer learning, by utilising networks trained on the larger dataset \mi and fine tune them to predict \mpartshort. Furthermore, we suggest a supervised domain adaption technique for jointly train the detection system and improve results. Hence, instead of tuning \mpartshort after the training is finished on \mi, we inject degrees of \mpartshort to \mi during training while evaluating the improvement on \mpartshort.

\subsection{Car Part Recognition}
\begin{figure}[t!]
 \begin{minipage}{\columnwidth} \centering 
  \includegraphics[width=\textwidth]{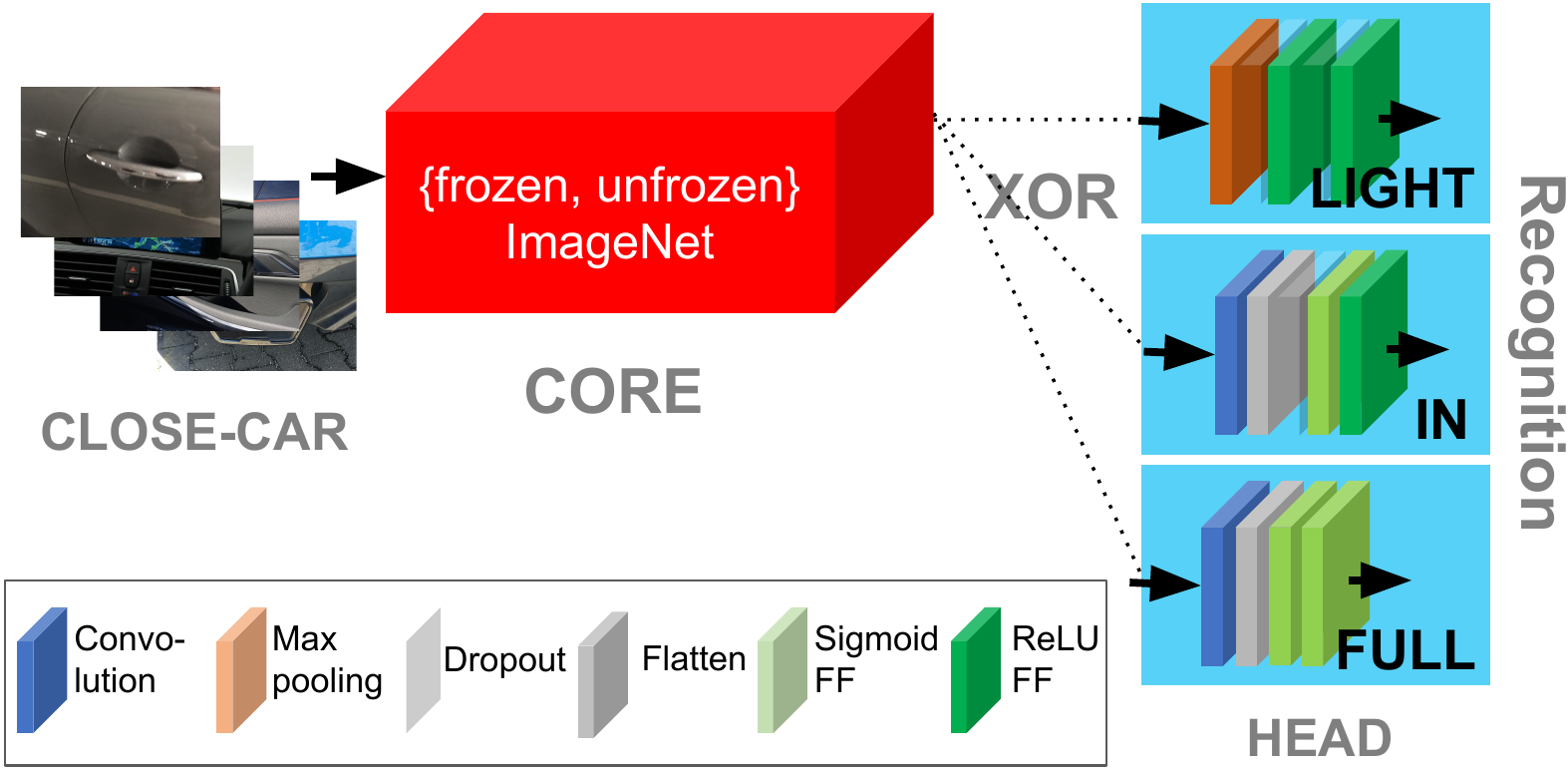} 
 \end{minipage}
  \caption{Proposed architectures from the data input to the prediction for the task of car part recognition.}
  \label{fig:recognition} 
\end{figure}

\begin{figure}[t!]
 \begin{minipage}{\columnwidth} \centering 
  \includegraphics[width=\textwidth]{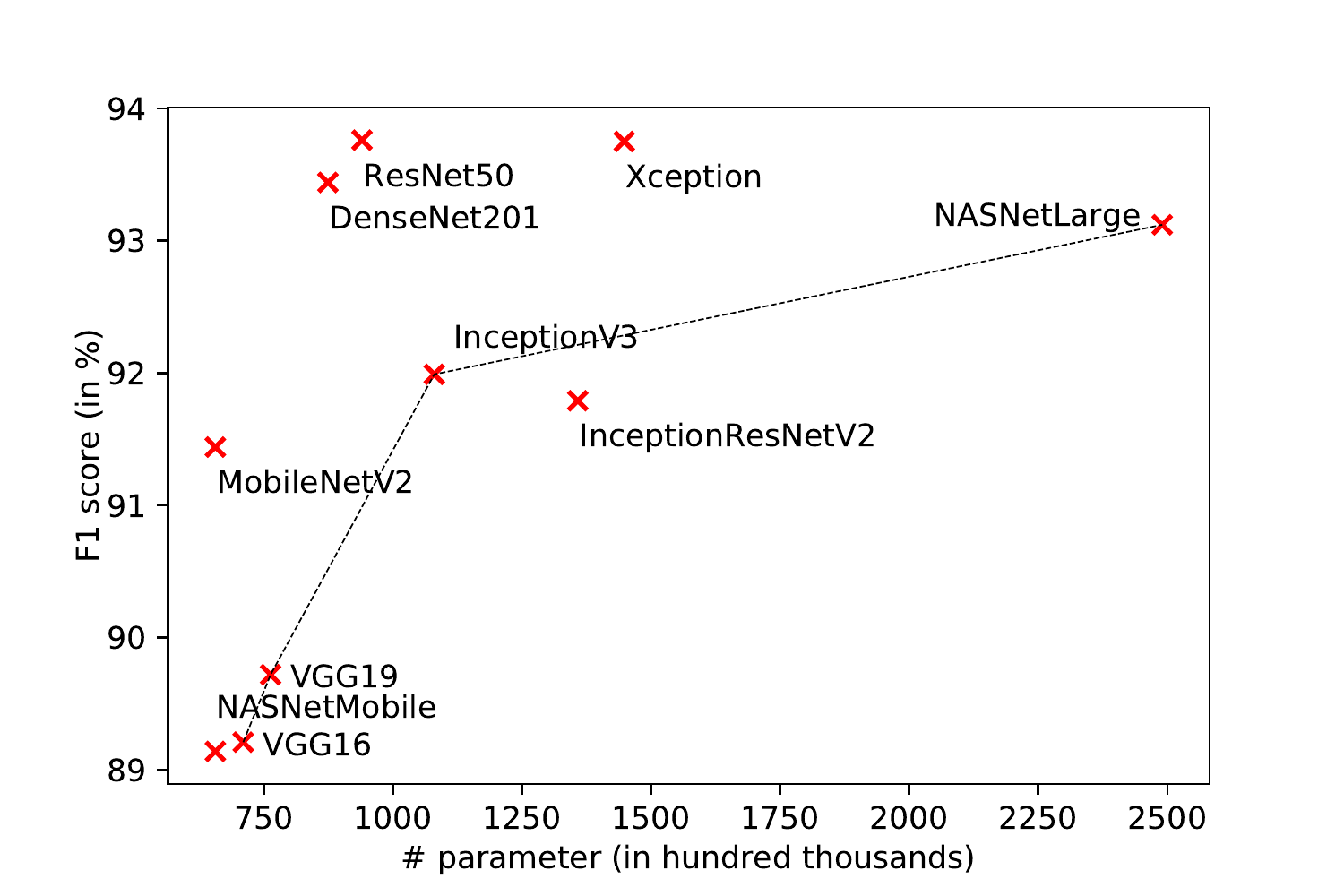} 
 \end{minipage}
  \caption{Number of parameters (in hundred thousands) of all fine-tuned ($FULL$) models compared to the achieved F1 score (combined) on the test set. }
  \label{fig:fullshotpara} 
\end{figure}

The recognition networks consist of a network base and a head. The base is initialised with weights trained on the ImageNet~\cite{deng2009imagenet} dataset of approximately 1.2 million images from 1\,000 categories.
The head, corresponding to the final (top) layers of the network, has a random initialisation.
We compare a static (functions as a feature extractor) and trainable base in combination with different heads as depicted in \Cref{fig:recognition}:
\textit{i) \textsc{light}}-weighted head on the top of the frozen base model, in which we apply max-pooling, followed by a 512 and 256 ReLU layer, using 50\,\% dropout for regulation; 
\textit{ii)} a parameter-intensive (\textit{\textsc{int}}) head, in which we add a trainable 2D convolutional layer with 2\,048 filters, a kernel size of 3x3, and valid padding, on the frozen base, followed by two fully-connected dense layers (1\,024 neurons with a sigmoid and 256 neurons with a ReLU activation function, respectively);
\textit{iii) \textsc{full}} has a trainable base topped with a head consisting of a 2D convolutional layer with 1\,024 filters, a kernel size of 3x3 followed by a 1\,024- and a 256-fully connected layer with a sigmoid activation function.

We also evaluated several other combinations and layer configurations, \eg varying number of neurons per layer. Given worse or very similar results, we decided to omit these for conciseness. The models are trained for up to 400 epochs. We counteracted adverse effects of class imbalance by taking the class weights in the loss functions into account.  
We run a hyperparameter search for a batch size = $\{16, 32, 64, 128\}$ applying an Adam optimiser with a learning rate = $\{0.1, 0.05, 0.01, 0.005, 0.001\}$. All experiments are executed on in- and outside downsized images\footnote{224 x 224: \dn, \mnm, \rn; 299 x 299: \irn, \iv, \xce; 331 x 331: \mnl}, reporting F1 score separately and combined. 

\subsection{Car Part Detection}
We train the backbone networks in a two step procedure: First, to smooth the training, only the last 3 layers are trained for 50 epochs, with a learning rate of $\alpha_1$; second, we unfreeze all layers and re-initialise the learning rate to $\alpha_2$, where $\alpha_2$ is a tenth of $\alpha_1$, \eg $\alpha_1 = 0.1$ and $\alpha_2 = 0.001$. 
The learning rate $\alpha_2$ is further reduced by the factor of 0.1 when the validation loss stagnates for 5 epochs and runs up to 200 epochs. To increase stability, we apply gradient clipping at 0.5 to the \squee. 
Similar to recognition, we weight the loss depending on the number of classes.
The input size is set to 416 x 416. 
Crafted using kmeans, the \squee and \darknet utilise nine anchors while the \dtiny utilise six. We use three thresholds to report the mAP under poor ($> 0.2$), moderate ($> 0.4$), and good ($> 0.5$) fit.

\section{RESULTS}\label{sec:results}

\subsection{Quantitative results}
\begin{table}[t!]
 	\caption{
 	    Recognition results of \close on the devel(opment) and test set reported \textbf{F1} in [\%] and the number of parameters in hundred thousands. 
      	Configuration ($Conf.$): \textsc{light}, \textsc{int}, \textsc{full}.
		}
	\label{tab:results_score}
\centering
  \resizebox{\columnwidth}{!}{
    \begin{tabular}{c|cccccc|cc}
    \toprule
    & \multicolumn{2}{c}{\textbf{outside}} & \multicolumn{2}{c}{\textbf{inside}} & \multicolumn{2}{c}{\textbf{combined}}  & \multicolumn{2}{c}{\textbf{\# parameter}} \\
     & dev & test& dev  & test & dev  & test  & trainable & frozen \\   
    \hline
    \hline
    $Conf.$  & \multicolumn{6}{c}{\textbf{\textsc{\dn}}}\\
    \hline
    \textsc{light} &67.37 &69.90 &55.63 &58.58 &63.94 &59.98 &11 &183\\
    \textsc{int} &88.88 &90.75 &77.00 &78.17 &77.82 &77.11 &880 &183\\
    \textsc{full} &97.13 & \underline{\textbf{93.08}} &87.35 &89.32 &94.63 &93.44 &874 &2\\
    \cline{2-9}
    & \multicolumn{6}{c}{\textbf{\textsc{\irn}}}\\
    \cline{2-9}
    \textsc{light} &68.46 &70.54 &48.50 &53.48 &63.46 &62.56 &12 &543\\
    \textsc{int} &68.97 &70.13 &68.73 &68.13 &78.26 &78.24 &1386 &543\\
    \textsc{full} &79.14 &80.33 &67.84 &67.29 &73.87 &85.76 &723 &0\\
    \cline{2-9}
    & \multicolumn{6}{c}{\textbf{\textsc{\iv}}}\\
    \cline{2-9}
    \textsc{light}	&79.25	&75.27	&62.42	&49.41	&66.73	&65.81	&11	&218\\
    \textsc{int} &86.93 &74.07 &65.14 &67.61 &79.1 &79.55 &1513 &218\\
    \textsc{full} &86.29 &78.19 &75.94 &79.21 &82.55 &87.56 &289 &0\\
    \cline{2-9}
    & \multicolumn{6}{c}{\textbf{\textsc{\mnl}}}\\
    \cline{2-9}
    \textsc{light} &82.13 &75.64 &51.64 &50.81 &65.02 &67.68 &28 &849\\
    \textsc{int} &91.32 &79.52 &66.83 &61.86 &71.65 &71.46 &3258 &849\\
    \textsc{full}	&99.45	&92.68	&87.98	&86.46	&87.4	&93.12	&2490	&1\\
    \cline{2-9}
    & \multicolumn{6}{c}{\textbf{\textsc{\rn}}}\\
    \cline{2-9}
    \textsc{light} &4.73 &2.97 &1.60 &2.13 &0.47 &0.63 &11 &235\\
    \textsc{int} &2.46 &2.60 &1.54 &1.20 &0.5 &0.49 &904 &235\\
    \textsc{full} &93.12 &90.20 &87.73 &89.51 &90.23 & \underline{\textbf{93.76}} &940 &0\\
    \cline{2-9}
    & \multicolumn{6}{c}{\textbf{\textsc{\vn}}}\\
    \cline{2-9}
    \textsc{light} &81.30 &83.11 &71.97 &74.03 &69.51 &69.59 &3 &200\\
    \textsc{int} &91.40 &87.53 &88.14 &77.85 &80.64 &78.81 &621 &200\\
    \textsc{full} &96.97 &91.39 &79.37 &78.50 &82.42 &89.72 &763 &0\\
    \cline{2-9}
    & \multicolumn{6}{c}{\textbf{\textsc{\xce}}}\\
    \cline{2-9}
    \textsc{light} &82.11 &77.49 &65.12 &72.06 &66.86 &68.9 &11 &208\\
    \textsc{int} &86.26 &88.96 &79.57 &75.28 &79.88 &80.52 &1722 &208\\
    \textsc{full} &97.02 &89.06 &84.52 &88.64 &90.35 &93.75 &1448 &0\\
    \hline
    \hline
    & \multicolumn{6}{c}{\textbf{\textsc{\mnv}}}\\
    \cline{2-9}
    \textsc{light} &75.58 &68.68 &56.72 &49.26 &60.12 &59.68 &7 &22\\
    \textsc{int} &79.27 &87.66 &73.16 &79.22 &77.39 &77.09 &762 &22\\
    \textsc{full} &93.57 &90.77 &87.00 & \underline{\textbf{92.60}} &84.39 &91.44 &656 &0\\
    \cline{2-9}
    & \multicolumn{6}{c}{\textbf{\textsc{\mnm}}}\\
    \cline{2-9}
    \textsc{light} &58.44 &69.88 &47.21 &56.40 &55.37 &53.11 &6 &42\\
    \textsc{int} &77.29 &77.64 &71.52 &72.04 &70.58 &65.01 &721 &42\\
    \textsc{full} &92.22 &91.21 &77.30 &87.96 &85.81 &89.14 &656 &0\\
    \bottomrule
    \end{tabular}
    }
\end{table}

\subsubsection{Recognition}
~\Cref{tab:results_score} depicts detailed results of car part recognition, demonstrating that models without frozen parameters (\textsc{full}) consistently yield the best results, except for \irn (inside). 
\rn achieved $93.76$\,\%, followed by \xce and \dn (combined). Both variants with a frozen network base using the comprehensive (\textsc{int}) and the simpler head (\textsc{light}), perform considerably worse. This indicates that fine-tuning the entire model and the head using a convolutional layer, instead of max-pooling, is worthwhile for the task. 

Another aspect of the training procedure are the number of weights to be learnt. ~\Cref{fig:fullshotpara} shows that, these three networks also have the best efficiency in terms of parameter usage compared to the other ones (indicated by the dotted line), while \irn clearly underperforms. Of the two models specifically developed for mobile applications, \mnv clearly outperforms \mnm and achieves also the best overall result in the detection of interior parts with almost identical numbers of parameters. 

\begin{table}[tp]
 	\caption{
 	    Detection results using a dataset for training (\textsc{T1}) with optional data injections in [\%] of the second training set (\textsc{T2}) to compare unsupervised domain adaption (T1 $\neq$ T2), joint learning (JL) and fine-tuning (FT); considering three levels of IoU fit for reporting \textbf{mAP} in [\%] on the development and test set (\textsc{dev./test}). 
		}
	\label{tab:results_det}
\centering
  \resizebox{\columnwidth}{!}{
    \begin{tabular}{ccccc|rr|rr|rr}
    \toprule
    \multicolumn{5}{c}{\textbf{data}} & \multicolumn{6}{c}{\textbf{IoU level on dev. / test}} \\
    T1 & M & [\%] & T2 & dev/test & \multicolumn{2}{c}{$> 0.2$} & \multicolumn{2}{c}{$> 0.4$}  & \multicolumn{2}{c}{$> 0.5$} \\   
    \hline
    \hline
    \multicolumn{10}{c}{\textbf{\textsc{\darknet}}}\\
    \hline 
    \textsc{part} & -- & -- & -- & \textsc{part} & 27.32 & 16.83 & 25.25 & 14.89 & 22.24 & 13.13  \\
        \textsc{part} & JL & $50$ & \textsc{mix} & \textsc{part} & 38.88 & 28.03 & 35.95 & 25.05 & 33.95 & 22.56 \\
    \textsc{part} & JL & $100$ & \textsc{mix} & \textsc{part} & \underline{\textbf{49.42}} & \underline{\textbf{41.07}} & 46.81 & 38.60 & 41.92 & 35.56  \\
    \textsc{part} & FT & $100$ & \textsc{mix} & \textsc{part} & 41.12 & 30.07 & 38.76 & 27.64 & 35.96 & 24.98  \\
    \textsc{mix} & -- & -- & -- & \textsc{part} & 26.15 & 17.39 & 24.81 & 15.89  & 22.71 & 14.46 \\
    \textsc{mix} & JL & $10$ & \textsc{part} & \textsc{part} & 26.69 & 22.24 & 24.70 & 20.71 & 23.13 & 19.06 \\
    \textsc{mix} & JL & $25$ & \textsc{part} & \textsc{part} & 33.60 & 23.16 & 31.23 & 20.74 & 29.37 & 18.09 \\
    \textsc{mix} & JL & $50$ & \textsc{part} & \textsc{part} & 43.01 & 29.89 & 41.16 & 27.65 & 36.37 & 25.01 \\
    \textsc{mix} & JL & $75$ & \textsc{part} & \textsc{part} & 45.60 & 34.89 & 43.13 & 32.14 & 39.73 & 29.74 \\
    \textsc{mix} & JL & $100$ & \textsc{part} & \textsc{part} & \underline{\textbf{49.42}} & \underline{\textbf{41.07}} & 46.81 & 38.60 & 41.92 & 35.56  \\
    \textsc{mix} & FT & $100$ & \textsc{part} & \textsc{part} & 42.10 & 29.62 & 39.37 & 27.46 & 35.06 & 23.80 \\ 
    \textsc{mix} & -- & -- & -- & \textsc{mix} & 59.43 & 58.20 & 58.32 & 56.66 & 56.30 & 54.60 \\
    \textsc{mix} & JL & $50$ & \textsc{part} & \textsc{mix} & 51.02 & 50.04 & 49.69 & 48.39 & 46.70 & 46.09 \\
    \textsc{mix} & JL & $100$ & \textsc{part} & \textsc{mix} & \underline{\textbf{65.28}} & \underline{\textbf{63.01}} & 63.83 & 61.22 & 61.89 & 59.10 \\
    \hline
    \multicolumn{10}{c}{\textbf{\textsc{\dtiny}}}\\
    \hline
    \textsc{mix} & JL & $100$ & \textsc{part} & \textsc{part} & 32.50 & 28.24 & 17.66 & 14.44 & 12.61 & 12.51 \\
    \textsc{mix} & -- & -- & -- & \textsc{mix} & 42.31 & 40.89 & 28.13 & 26.43 & 25.61 & 24.41 \\
    \hline
    \multicolumn{10}{c}{\textbf{\textsc{\squee}}}\\
    \hline
    \textsc{mix} & JL & $100$ & \textsc{part} & \textsc{part}  & 34.74 & 22.99 & 33.19 & 21.00 & 29.98 & 19.07 \\
    \textsc{mix} & -- & -- & -- & \textsc{mix} & 48.11 & 46.03 & 46.74 & 44.14 & 44.24 & 42.29\\
    \bottomrule
    \end{tabular}
    }
\end{table}

\subsubsection{Detection}
For car part detection, \darknet clearly shows the strongest performance of all network backbones, regardless of whether the dataset is predicted without (dev/test = \mi) or with human obstruction (dev/test = \mpartshort). For example, on the \mi test set (IoU $>$ 0.5), it achieves a mAP  of $54.60$\,\%, followed by $42.29$ for \squee and $24.41$ for \dtiny (\cf \Cref{tab:results_score}). It is evident that even using the best architecture \darknet, training on the smaller data set with human interactions alone (\mpartshort to \mpartshort) is very difficult, resulting in $27.32$ on the development set and $16.83$ on the test set at IoU $>$ 0.2 level. A purely unsupervised in-domain adaption from \mi (=T1) to \mpartshort (=dev/test) without fine-tuning or injecting training data of the target domain results in almost the same, low prediction performance ($26.15$/ $17.39$\,\% at an IoU level $>$ 0.2). Both results show the need to adopt more advanced training techniques.

Using the trained \mi weights for fine-tuning (FT) with all (100) \mpartshort data, improve results to up to $42.10$\,\% on the development and $29.62$\,\% on the test set (IoU $>$ 0.2). Interestingly, training in the opposite direction leads to very similar, albeit slightly worse, results. This could be an indicator that the influence of the people from \mpartshort is retained even with additional fine-tuning of the representations, but that the general object understanding is greatly improved due to more data. 
We also experimented with with varying learning rates and isolated fine-tuning on certain layers of the model (\eg the first x layers), which did not improve results further. 

The joint training domain adaption approach (JL) gradually improves results on \mpartshort (dev/test) when more and more ($10$\%-$100$\%) data are injected from \mpartshort (T2) to \mi (T1), until both training partitions are used entirely ($100$). Already $50$ \% injected from \mpartshort performs better than fine-tuning ($43.01$/ $29.89$ \% at $0.2$ IoU). However, when injecting the other way around (\mi to \mpartshort) the performance is slightly worse. $100$ \% achieves an mAP of $41.07$\,\% (at $0.2$ IoU) on test, which is an improvement of $11.45$\,\% compared to the fine-tuned approach.
For the prediction of \mi, we can also improve the performance by almost $5$\,\% compared to the standard training procedure. However, when only $50$\,\% of the \mpartshort data is injected, the results are worse, suggesting that a certain amount of data is required to learn from the new diversity of another domain and not just inject additional, unfavorable noise to an already stable training.

\begin{figure}[t!] 
\begin{minipage}{\columnwidth}\centering 
  \includegraphics[width=.8\textwidth]{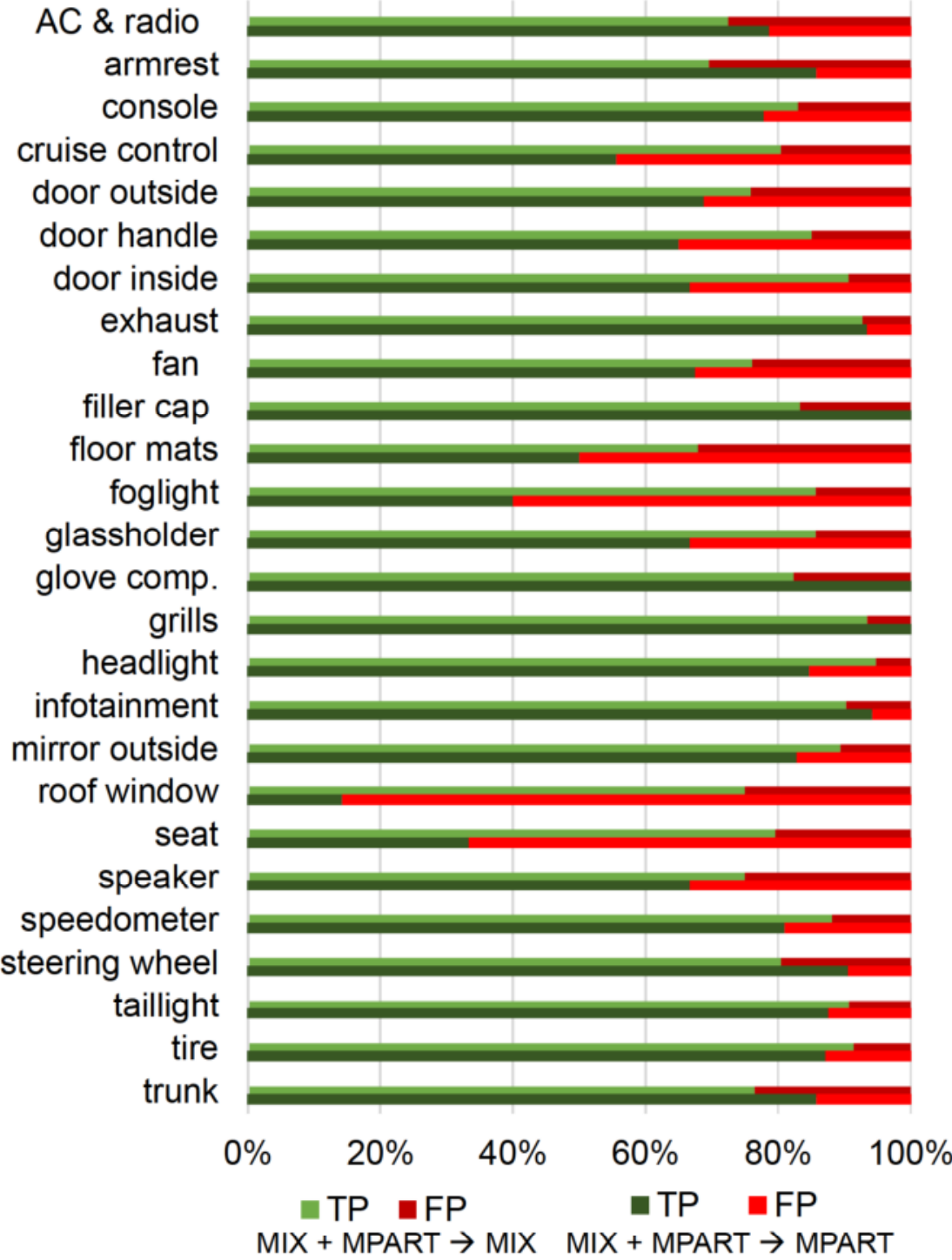} 
 \end{minipage}
  \caption{A relative comparison of the joint-learning capabilities on both datasets illustrating the True Positive (TP) and False Positives (FP) predictions per class of the \textsc{mix} + 100 \% \textsc{PART} $\rightarrow$ \textsc{mix} and \textsc{mix} + 100 \% \textsc{PART} $\rightarrow$ \textsc{PART} at .2 level.}
  \label{fig:fp_tp_distribution} 
  \vspace{-8pt}
\end{figure}

\Cref{fig:fp_tp_distribution} illustrates all predictions of the best models as a relative comparison of the True Positive (TP) and False Positives (FP) predictions for the two training modes. For predictions on the jointly trained system (\mi + \mpartshort predicting \mpartshort), more than two-thirds of the classes have more than 80\,\% TPs. In contrast, the jointly trained system (\mi + \mpartshort predicting \mpartshort) predicts some classes, such as roof window, seat, and foglight, more wrong than right. Both achieve above average results when either the class has a large number of labels (\eg tyre, headlight) or the objects have very distinctive features (\eg exhaust, front grill).


\begin{figure*}[t!] 
 \centering
  \includegraphics[width=0.8\textwidth]{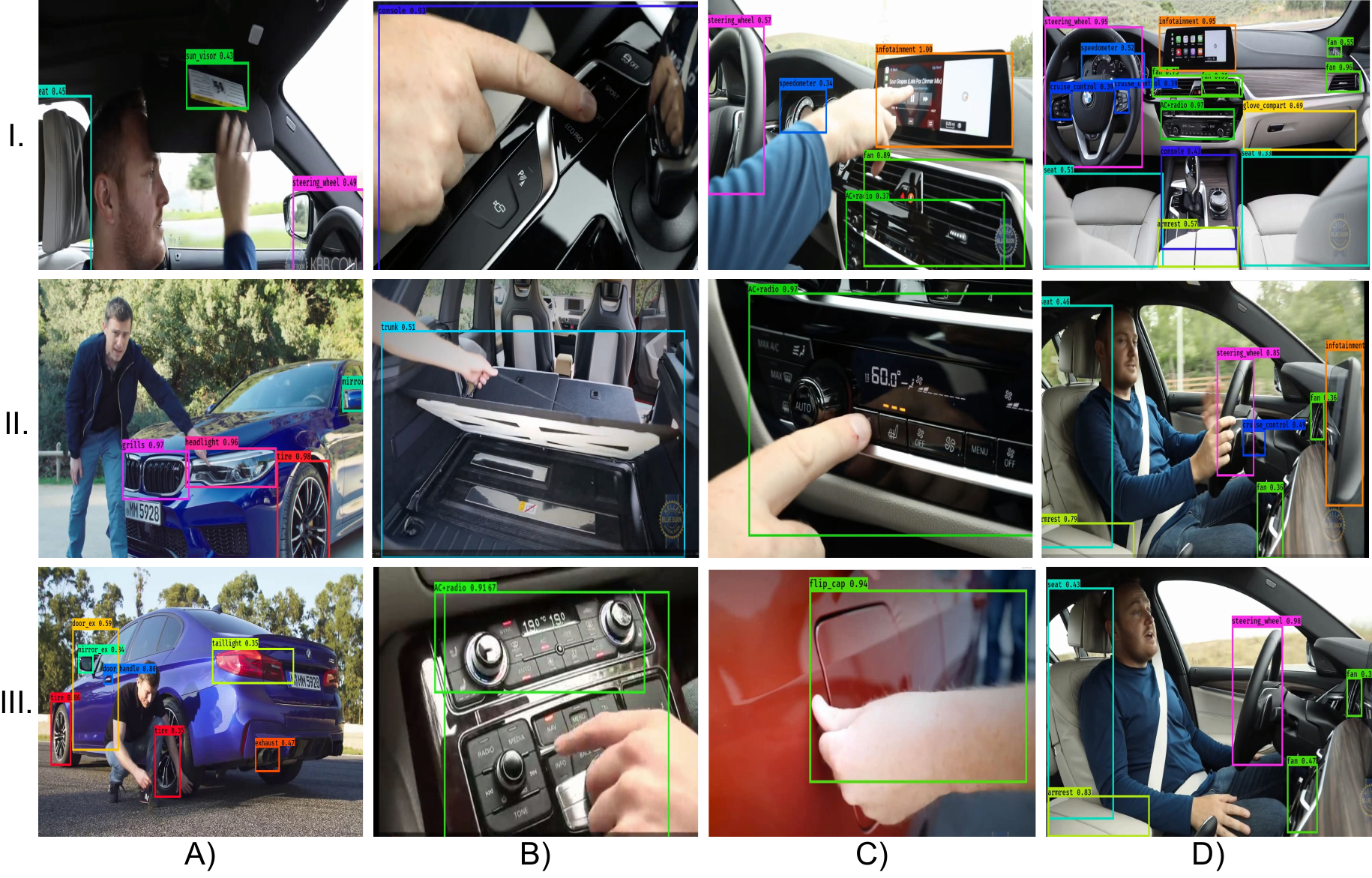} 

  \caption{Example frames MuSe-CaR: A) typical interaction between a partially obscured human and car parts (I. sun visor, II. headlight, III. tire); B) + C) hand interactions and gestures towards objects; and D) I. area separation of the car interior II. driving (hands on wheel) III. semi-autonomous driving (hands off wheel).}
  \label{fig:yoloinsights} 
\end{figure*}

\subsection{Qualitative results}
\subsubsection{Recognition}
~\Cref{fig:fullshot} depicts examples of parts incorrectly recognised by the best-performing architecture. 
A and B are both predicted as floor mats, probably due to patterns on the fabric below the A/C (A) and backrest (B), indicating a sensitivity to distinctive patterns. Certain classes in less common variations, such as the open glove compartments in C and D, appear to be more vulnerable to variations in lighting. 
Similarly, images with high contrast combined with reflections, such as the tinted sun roof of a white car (E) and the reflection of a metal exhaust in the dark (F), are more prone to confusion. 

\subsubsection{Detection}
\begin{figure*}[tp] 
  \includegraphics[width=\textwidth]{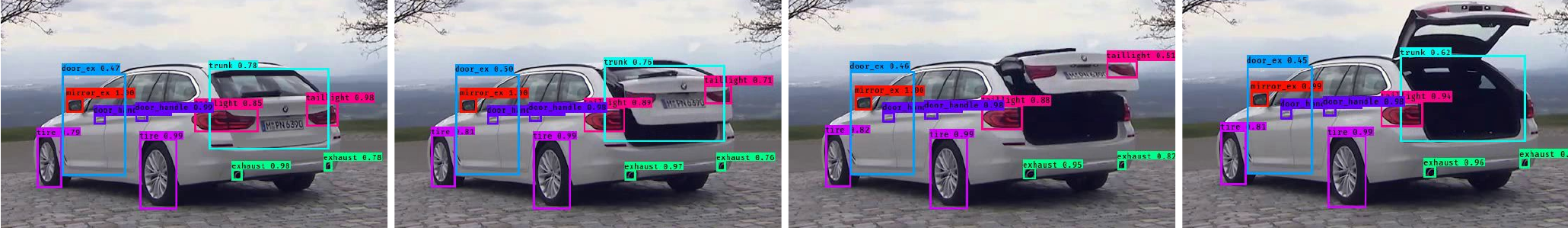} 
  \caption{Example frames demonstrating the 'loss' of a moving object, the trunk door in the third frame, on a not considered video of  MuSe-CAR using the \darknet backbone architecture trained on non-consecutive frames.}
  \label{fig:trunk} 
\end{figure*}

For the qualitative analysis, we do not test the best detection network on individual images, but on videos that are included in MuSe-CaR but not in \mix.
We found that the models have a high detection rate in distance shots, for example, door handles on distant, approaching vehicles that are hardly visible to the human eye. We attribute this to the learnt context. For instance, a door handle is always located at the same position of the door and the relative location only depends on the camera perspective.
Our transfer and fine-tuning approach drastically improved the robustness when people interact with an object, especially in the case of minor occlusions due to finger pointing or gestures, such as B and C in~\Cref{fig:yoloinsights}. Greatly improved, although still with limitations, is the detection of objects that are gripped (A-I.: the bounding box around the sun visor is reduced in size due to the hand-grip) or are obscured by human body parts (A-III. the rear door is largely obscured and not detected) in joint usage of both datasets. In one limitation, used on consecutive video frames the models temporarily 'lose' objects with moving parts such as when a door or the trunk is opened, as in the example in~\Cref{fig:trunk}). Additional distinction between classes (open, closed) and the extraction and annotation of similar frames should overcome this issue.


\section{APPLICATIONS}\label{sec:app} 
A reliable detection of objects inside vehicles has many potential applications. The \textbf{interior} becomes increasingly important with the evolution of autonomous driving and the growing number of customer functions dedicated to communication and entertainment, in addition to future use cases, such as mobile working by occupants as the car drives to its destination. Object and passenger detection forms a basic component for a model of the car cabin. One example is the monitoring of the driver for unexpected take over scenarios (\cf~\Cref{fig:yoloinsights}, D) II. to III.) in semi-autonomous driving \cite{9210541,9334426}. Another is gesture control, an intuitive form of user interaction, that can be implemented by localising individual fingers and their relation to objects within the vehicle \cite{9151384}. 
Which parts of the car a passenger is pointing towards or interacting with is important information for an advanced intelligent vehicle assistant (\cf~\Cref{fig:yoloinsights}, B and C). Such systems depend on context to develop an understanding of the user's intent~\cite{vogel2018emotion, 9220832}. 
Similarly, \gocart was already successfully applied for the task of driver gaze prediction in `in-the-wild' environments \cite{stappen2020xaware}. The context-aware (carbine-face) feature fusion allowed implicit calibration of the face to objects. This is a crucial advantage when the position of the camera is constantly changing or when detection systems should be transferred without additional training from one car model to another, which mostly also implies a different distance to the subject. Explicitly understanding the environment makes these systems more generalisable compared to purely human/face-centred approaches, since some parts of a car \eg the upper anchorage of the safety belt are visible independently of make or model. They can elevate the driving experience through smart assistance upon recognising driver distraction or stress. By learning the passenger's preferences, these models can also provide personalised user interfaces or change the interior configuration for increased comfort. 
 Finally, the contextual input provided by computer vision algorithms and other sensor sources can help intelligent assistants anticipate the passengers' intentions and act proactively \cite{vogel2018emotion,6629497}.
 
Applications of part detection \textbf{outside} the vehicle exist in autonomous driving, \eg collision avoidance of opening parts on nearby vehicles. In case of crashes, part localisation may help reconstruct the accidents for insurance claims. Signs of wear may also be detected~\cite{Balitskii.2019}, allowing timely maintenance.
Automotive production can also benefit from part recognition and detection as well, with applications in oncoming humanoid robot generations, process monitoring and quality control. Use cases include verifying manufacturing steps, for example, via virtual inspection~\cite{Luckow.2016b}, or generating models that simulate entire factories~\cite{Petschnigg.2020}.

Generally, the ability to detect specific parts of vehicles may also be useful for sales and marketing, where the use of multiple modalities is a promising direction for sentiment analysis~\cite{Karas.2020}. Incorporating visual information into multimodal approaches is a valuable step towards understanding videos reviewing vehicles, a large number of which are available online~\cite{stappen2020muse}.

\section{CONCLUSION}
\label{sec:conc} 
As a potential cornerstone for a diverse set of deep learning tasks in the automobile domain, we develop a generic, optical car part recognition and detection system for realistic conditions.
To do so, we introduced three new datasets, each the biggest of its kind, containing images with varying make and models in illuminated and obstructed environments. These datasets allowed us to develop various computer vision models and empirically evaluate out- and in-domain transfer and joint learning concepts. On classification dataset, our recognition systems achieved an F1 score of $93.67$\,\% (test set). For detection, both of our jointly trained systems performed better compared to the standard training procedure and transfer learning. It improved results (absolute) by more than 5\,\% to $65.28$\,\% and $63.01$\,\% mAP on the development and test set of \mix. Similarly, results improved by $10$\% from $29.62$\,\% to $41.07$\,\% mAP on the human-car interaction dataset  (test set) compared to fine-tuning. 
In addition, we hope that our extensive result description and proposed applications provide useful guidance for practitioners and researchers to come closer to our goal of a fully generic system. To support this goal, we released our trained models and the human-car interaction dataset for future research and benchmarking. Furthermore, we plan to examine human-vehicle interactions within this setting more closely and in additional environments. 


\vspace{-0.8em}
\section{ACKNOWLEDGMENTS}
We thank the BMW Group for the provision of the data and computation resources. 

\ifCLASSOPTIONcaptionsoff
  \newpage
\fi



{
\bibliographystyle{IEEEtran} 
\bibliography{Main.bib}
}

\begin{IEEEbiography}[{\includegraphics[width=1in,height=1.25in,clip,keepaspectratio]{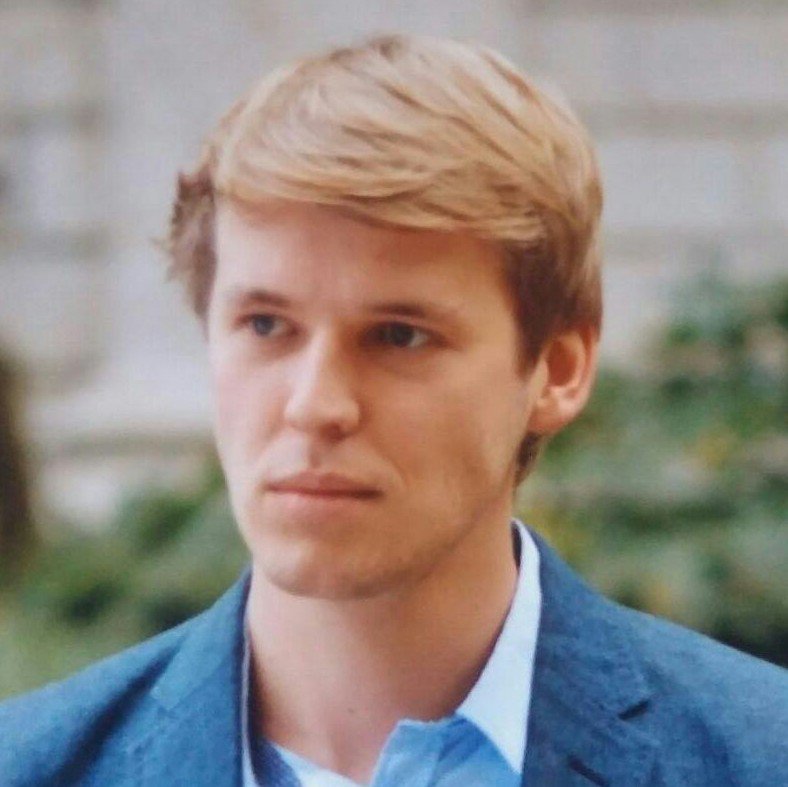}}]{Lukas Stappen}
received his Master of Science in Data Science with distinction from King's College London in 2017. He then joined
the group for Machine Learning in Health Informatics. Currently, he is a PhD candidate at the Chair for Embedded Intelligence for Health Care and Wellbeing, University of Augsburg, Germany, and a PhD Fellow of the BMW Group. His research interests include affective computing, multimodal sentiment analysis, and multimodal/cross-modal representation learning with a core focus 
on `in-the-wild' environments. He received several student academic excellence scholarships as well as the best paper award of the IEEE MMSP 2020, and served as general chair at the MuSe 2020 Workshop (ACM Multimedia).
\end{IEEEbiography}


\begin{IEEEbiography}[{\includegraphics[width=1in,height=1.25in,clip,keepaspectratio]{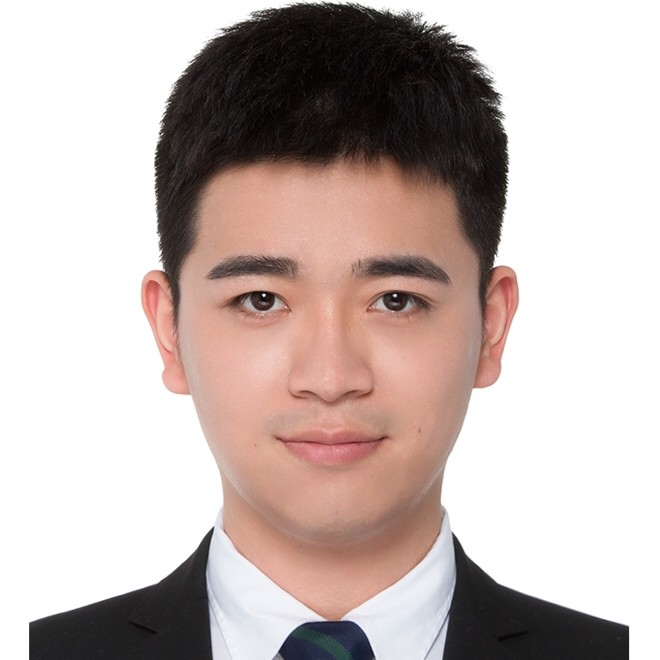}}]{Xinchen Du}
received his B.S. degree in electrical engineering from Tongji University, Shanghai, China, in 2016. He recently completed a
M.Sc. degree in electrical engineering from Technical University of Munich, Germany. His research interests include robotics, computer vision and natural language processing.
\end{IEEEbiography}

\begin{IEEEbiography}[{\includegraphics[width=1in,height=1.25in,clip,keepaspectratio]{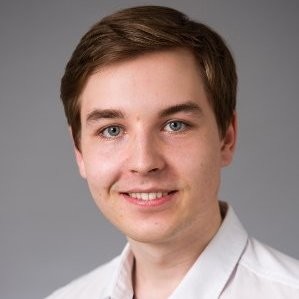}}]{Vincent Karas}
received his Bachelor's degree in Engineering Science from Technische Universität München (TUM) in 2014. He then pursued a Master's degree in Medical Technology and Engineering and TUM, graduating in 2018. Currently, he is a PhD candidate at the Chair for Embedded Intelligence for Health Care and Wellbeing, University of Augsburg, and a member of the doctoral candidate program at BMW AG. His research interests include affective computing in the automotive environment, multimodal emotion recognition in the wild, transfer learning, representation learning and estimating affect from physiological signals.
\end{IEEEbiography}

\begin{IEEEbiography}[{\includegraphics[width=1in,height=1.25in,clip,keepaspectratio]{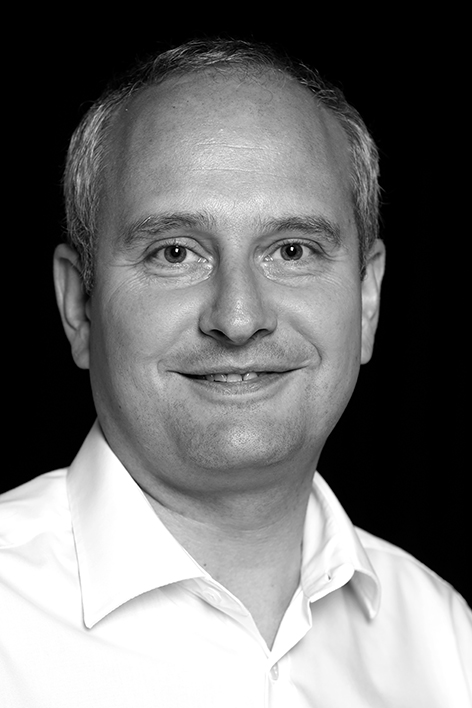}}]{Stefan Müller}
is an senior engineer at BMW Group, Munich, Germany and is responsible for the big data analytics team in the quality management department. Before joining BMW Group he received a diploma in mechanical engineering in 2001 and a Ph.D. in the field of production technologies from Technical University Munich (TUM) in 2007. His research interest are in the field of applied analytics and AI technologies in the automotive industry.
\end{IEEEbiography}


\begin{IEEEbiography}
    [{\includegraphics[width=1in,height=1.25in,clip,keepaspectratio]{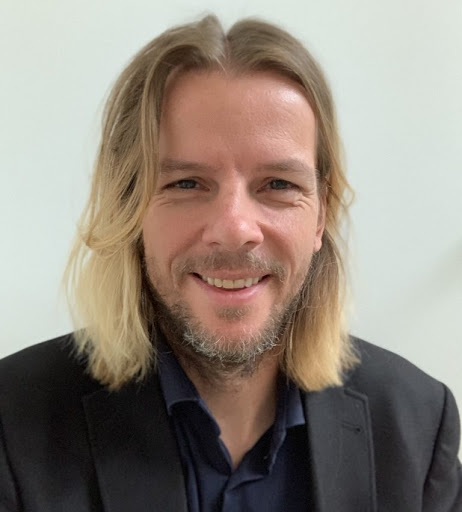}}]{Björn Schuller}
received his diploma, doctoral degree, habilitation, and Adjunct Teaching Professor in Machine Intelligence and Signal Processing all in EE/IT from TUM in Munich/GER. He is Full Professor of Artificial Intelligence and the Head of GLAM at Imperial College London/UK, Full Professor and Chair of Embedded Intelligence for Health Care and Wellbeing at the University of Augsburg/GER, and permanent Visiting Professor at HIT/China amongst other Professorships and Affiliations. Previous stays include Full Professor at the University of Passau/GER, and Researcher at Joanneum Research in Graz/Austria, and the CNRS-LIMSI in Orsay/France. He is a Fellow of the IEEE and Golden Core Awardee of the IEEE Computer Society, Fellow of the ISCA, President-Emeritus of the AAAC, and Senior Member of the ACM. He (co-)authored 900+ publications (30k+ citations, h-index=83), is Field Chief Editor of Frontiers in Digital Health and was Editor in Chief of the IEEE Transactions on Affective Computing amongst manifold further commitments and service to the community. His 30+ awards include having been honoured as one of 40 extraordinary scientists under the age of 40 by the WEF in 2015. 
\end{IEEEbiography}

\end{document}